# Algorithms for Approximate Minimization of the Difference Between Submodular Functions, with Applications

**Rishabh Iyer**
Dept. of Electrical Engineering
University of Washington
Seattle, WA-98175, USA

**Jeff Bilmes**
Dept. of Electrical Engineering
University of Washington
Seattle, WA-98175, USA

## Abstract

We extend the work of Narasimhan and Bilmes [30] for minimizing set functions representable as a difference between submodular functions. Similar to [30], our new algorithms are guaranteed to monotonically reduce the objective function at every step. We empirically and theoretically show that the per-iteration cost of our algorithms is much less than [30], and our algorithms can be used to efficiently minimize a difference between submodular functions under various combinatorial constraints, a problem not previously addressed. We provide computational bounds and a hardness result on the multiplicative inapproximability of minimizing the difference between submodular functions. We show, however, that it is possible to give worst-case additive bounds by providing a polynomial time computable lower-bound on the minima. Finally we show how a number of machine learning problems can be modeled as minimizing the difference between submodular functions. We experimentally show the validity of our algorithms by testing them on the problem of feature selection with submodular cost features.

## 1 Introduction

Discrete optimization is important to many areas of machine learning and recently an ever growing number of problems have been shown to be expressible as submodular function minimization or maximization (e.g., [19, 23, 25, 28, 27, 29]). The class of submodular functions is indeed special since submodular function minimization is known to be polynomial time, while submodular maximization, although NP complete, admits constant factor approximation algorithms. Let $V = \{1, 2, \cdots, n\}$ refer a ground set, then $f : 2^V \to \mathbb{R}$ is said to be submodular if for sets $S, T \subseteq V$, $f(S)+f(T) \geq f(S \cup T)+f(S \cap T)$ (see [10] for details on submodular, supermodular, and modular functions). Submodular functions have a diminishing returns property, wherein the gain of an element in the context of bigger set is lesser than the gain of that element in the context of a smaller subset. This property occurs naturally in many applications in machine learning, computer vision, economics, operations research, etc.

In this paper, we address the following problem. Given two submodular functions $f$ and $g$, and define $v(X) \triangleq f(X) - g(X)$, solve the following optimization problem:

$$\min_{X \subseteq V}[f(X) - g(X)] \equiv \min_{X \subseteq V}[v(X)]. \quad (1)$$

A number of machine learning problems involve minimization over a difference between submodular functions. The following are some examples:

- **Sensor placement with submodular costs:** The problem of choosing sensor locations $A$ from a given set of possible locations $V$ can be modeled [23, 24] by maximizing the mutual information between the chosen variables $A$ and the unchosen set $V \setminus A$ (i.e., $f(A) = I(X_A; X_{V \setminus A})$). Alternatively, we may wish to maximize the mutual information between a set of chosen sensors $X_A$ and a fixed quantity of interest $C$ (i.e., $f(A) = I(X_A; C)$) under the assumption that the set of features $X_A$ are conditionally independent given $C$ [23]. These objectives are submodular and thus the problem becomes maximizing a submodular function subject to a cardinality constraint. Often, however, there are costs $c(A)$ associated with the locations that naturally have a diminishing returns property. For example, there is typically a discount when purchasing sensors in bulk. Moreover, there may be diminished cost for placing a sensor in a particular location given placement in certain other locations (e.g., the additional equipment needed to install a sensor in, say, a precarious environment could be re-used for multiple sensor installations in like environments). Hence, along with maximizing mutual information, we also want to simultaneously minimize the cost and this problem can be addressed by minimizing the difference between submodular functions $f(A) - \lambda c(A)$ for tradeoff parameter $\lambda$.

- **Discriminatively structured graphical models and neural computation:** An application suggested in [30] and the initial motivation for this problem is to optimize the EAR criterion to produce a discriminatively structured graphical model. EAR is basically a difference between two mutual information functions (i.e., a difference between submodular functions). [30] shows how classifiers based on discriminative structure using EAR can significantly outperform classifiers based on generative graphical models. Note also that the EAR measure is the same as "synergy" in a neural code [2], widely used in neuroscience.

- **Feature selection:** Given a set of features $X_1, X_2, \cdots, X_{|V|}$, the feature selection problem is to find a small subset of features $X_A$ that work well when used in a pattern classifier. This problem can be modeled as maximizing the mutual information $I(X_A; C)$ where $C$ is the class. Note that $I(X_A; C) = H(X_A) - H(X_A|C)$ is always a difference between submodular functions. Under the naïve Bayes model, this function is submodular [23]. It is not submodular under general classifier models such as support vector machines (SVMs) or neural networks. Certain features, moreover, might be cheaper to use given that others are already being computed. For example, if a subset $S_i \subseteq V$ of the features for a particular information source $i$ are spectral in nature, then once a particular $v \in S_i$ is chosen, the remaining features $S_i \setminus \{v\}$ may be relatively inexpensive to compute, due to grouped computational strategies such as the fast Fourier transform. Therefore, it might be more appropriate to use a submodular cost model $c(A)$. One such cost model might be $c(A) = \sum_i \sqrt{m(A \cap S_i)}$ where $m(j)$ would be the cost of computing feature $j$. Another might be $c(A) = \sum_i c_i \min(|A \cap S_i|, 1)$ where $c_i$ is the cost of source $i$. Both offer diminishing cost for choosing features from the same information source. Such a cost model could be useful even under the naïve Bayes model, where $I(X_A; C)$ is submodular. Feature selection becomes a problem of maximizing $I(X_A; C) - \lambda c(A) = H(X_A) - [H(X_A|C) + \lambda c(A)]$, the difference between two submodular functions.

- **Probabilistic Inference:** We are given a distribution $p(x) \propto \exp(-v(x))$ where $x \in \{0,1\}^n$ and $v$ is a pseudo-Boolean function [1]. It is desirable to compute $\mathrm{argmax}_{x \in \{0,1\}^n} p(x)$ which means minimizing $v(x)$ over $x$, the most-probable explanation (MPE) problem [33]. If $p$ factors with respect to a graphical model of tree-width $k$, then $v(x) = \sum_i v_i(x_C)$ where $C_i$ is a bundle of indices such that $|C| \leq k+1$ and the sets $\{C_i\}_i$ form a junction tree, and it might be possible to solve inference using dynamic programming. If $k$ is large and/or if hypertree factorization does not hold, then approximate inference is typically used [38]. On the other hand, defining $x(X) = \{x \in \{0,1\}^n : x_i = 1 \text{ whenever } i \in X\}$, if the set function $\bar{v}(X) = v(x(X))$ is submodular, then even if $p$ has large tree-width, the MPE problem can be solved exactly in polynomial time [17]. This, in fact, is the basis behind inference in many computer vision models where $v$ is often not only submodular but also has limited sized $|C_i|$. For example, for submodular $v$ and if $|C_i| \leq 2$ then graph-cuts can solve the MPE problem extremely rapidly [22] and even some cases with $v$ non-submodular [21]. An important challenge is to consider non-submodular $v$ that can be minimized efficiently and for which there are approximation guarantees, a problem recently addressed in [18]. On the other hand, if $v$ can be expressed as a difference between two submodular functions (which it can, see Lemma 3.1), or if such a decomposition can be computed (which it sometimes can, see Lemma 3.2), then a procedure to minimize the difference between two submodular functions offers new ways to solve probabilistic inference.

Previously, Narasimhan and Bilmes [30] proposed an algorithm inspired by the convex-concave procedure [39] to address Equation (1). This algorithm iteratively minimizes a submodular function by replacing the second submodular function $g$ by it's modular lower bound. They also show that any set function can be expressed as a difference between two submodular functions and hence every set function optimization problem can be reduced to minimizing a difference between submodular functions. They show that this process converges to a local minima, however the convergence rate is left as an open question.

In this paper, we first describe tight modular bounds on submodular functions in Section 2, including lower bounds based on points in the base polytope as used in [30], and recent upper bounds first described in a result in [15]. In section 2.2, we describe the submodular-supermodular procedure proposed in [30]. We further provide a constructive procedure for finding the submodular functions $f$ and $g$ for any arbitrary set function $v$. Although our construction is NP hard in general, we show how for certain classes of set functions $v$, it is possible to find the decompositions $f$ and $g$ in polynomial time. In Section 4, we propose two new algorithms both of which are guaranteed to monotonically reduce the objective at every iteration and which converge to a local minima. Further we note that the per-iteration cost of our algorithms is in general much less than [30], and empirically verify that our algorithms are orders of magnitude faster on real data. We show that, unlike in [30], our algorithms can be extended to easily optimize equation (1) under cardinality, knapsack, and matroid constraints. Moreover, one of our algorithms can actually handle complex combinatorial constraints, such as spanning trees, matchings, cuts, etc. Further in Section 5, we give a hardness result that there does not exist any polynomial time algorithm with any polynomial time multiplicative approximation guarantees unless P=NP, even when it is easy to find or when we are given the decomposition $f$ and $g$, thus justifying the need for heuristic methods to solve this problem. We show, however, that it is possible to get additive bounds by showing polynomial time computable upper and lower bound on the optima. We also provide computational bounds for all our algorithms (including the submodular-supermodular procedure), a problem left open in [30].

Finally we perform a number of experiments on the feature selection problem under various cost models, and show how our algorithms used to maximize the mutual information perform better than greedy selection (which would be near optimal under the naïve Bayes assumptions) and with less cost.

## 2 Modular Upper and Lower bounds

The Taylor series approximation of a convex function provides a natural way of providing lower bounds on such a function. In particular the first order Taylor series approximation of a convex function is a lower bound on the function, and is linear in $x$ for a given $y$ and hence given a convex function $\phi$, we have:

$$\phi(x) \geq \phi(y) + \langle \nabla \phi(y), x - y \rangle. \qquad (2)$$

Surprisingly, any submodular function has both a tight lower [6] **and** upper bound [15], unlike strict convexity where there is only a tight first order lower bound.

### 2.1 Modular Lower Bounds

Recall that for submodular function $f$, the submodular polymatroid, base polytope and the sub-differential with respect to a set $Y$ [10] are respectively:

$$\mathcal{P}_f = \{x : x(S) \leq f(S), \forall S \subseteq V\} \qquad (3)$$
$$\mathcal{B}_f = \mathcal{P}_f \cap \{x : x(V) = f(V)\} \qquad (4)$$
$$\partial f(Y) = \{y \in \mathbb{R}^V : \forall X \subseteq V, f(Y) - y(Y) \leq f(X) - y(X)\}$$

The extreme points of this sub-differential are easy to find and characterize, and can be obtained from a greedy algorithm ([6, 10]) as follows:

**Theorem 2.1.** *([10], Theorem 6.11) A point $y$ is an extreme point of $\partial f(Y)$, iff there exists a chain $\emptyset = S_0 \subset S_1 \subset \cdots \subset S_n$ with $Y = S_j$ for some $j$, such that $y(S_i \setminus S_{i-1}) = y(S_i) - y(S_{i-1}) = f(S_i) - f(S_{i-1}).$*

Let $\sigma$ be a permutation of $V$ and define $S_i^\sigma = \{\sigma(1), \sigma(2), \ldots, \sigma(i)\}$ as $\sigma$'s chain containing $Y$, meaning $S_{|Y|}^\sigma = Y$ (we say that $\sigma$'s chain <u>contains</u> $Y$). Then we can define a sub-gradient $h_Y^f$ corresponding to $f$ as:

$$h_{Y,\sigma}^f(\sigma(i)) = \begin{cases} f(S_1^\sigma) & \text{if } i = 1 \\ f(S_i^\sigma) - f(S_{i-1}^\sigma) & \text{otherwise} \end{cases}.$$

We get a modular lower bound of $f$ as follows:

$$h_{Y,\sigma}^f(X) \leq f(X), \forall X \subseteq V, \text{ and } \forall i, h_{Y,\sigma}^f(S_i^\sigma) = f(S_i^\sigma),$$

which is parameterized by a set $Y$ and a permutation $\sigma$. Note $h(X) = \sum_{i \in X} h(i)$, and $h_{Y,\sigma}^f(Y) = f(Y)$. Observe the similarity to convex functions, where a linear lower bound is parameterized by a vector $y$.

### 2.2 Modular Upper Bounds

For $f$ submodular, [31] established the following:

$$f(Y) \leq f(X) - \sum_{j \in X \setminus Y} f(j|X \setminus j) + \sum_{j \in Y \setminus X} f(j|X \cap Y),$$
$$f(Y) \leq f(X) - \sum_{j \in X \setminus Y} f(j|(X \cup Y) \setminus j) + \sum_{j \in Y \setminus X} f(j|X)$$

Note that $f(A|B) \triangleq f(A \cup B) - f(B)$ is the gain of adding $A$ in the context of $B$. These upper bounds in fact characterize submodular functions, in that a function $f$ is a submodular function *iff* it follows either of the above bounds. Using the above, two tight modular upper bounds ([15]) can be defined as follows:

$$f(Y) \leq m_{X,1}^f(Y) \triangleq f(X) - \sum_{j \in X \setminus Y} f(j|X \setminus j) + \sum_{j \in Y \setminus X} f(j|\emptyset),$$
$$f(Y) \leq m_{X,2}^f(Y) \triangleq f(X) - \sum_{j \in X \setminus Y} f(j|V \setminus j) + \sum_{j \in Y \setminus X} f(j|X).$$

Hence, this yields two tight (at set $X$) modular upper bounds $m_{X,1}^f, m_{X,2}^f$ for any submodular function $f$. For briefness, when referring either one we use $m_X^f$.

## 3 Submodular-Supermodular Procedure

We now review the submodular-supermodular procedure [30] to minimize functions expressible as a difference between submodular functions (henceforth called DS functions). Interestingly, any set function

**Algorithm 1** The submodular-supermodular (SubSup) procedure [30]

1: $X^0 = \emptyset$ ; $t \leftarrow 0$ ;
2: **while** not converged (i.e., $(X^{t+1} \neq X^t)$) **do**
3:   Randomly choose a permutation $\sigma^t$ whose chain contains the set $X^t$.
4:   $X^{t+1} := \arg\min_X f(X) - h^g_{X^t, \sigma^t}(X)$
5:   $t \leftarrow t + 1$
6: **end while**

The submodular supermodular (SubSup) procedure is given in Algorithm 1. At every step of the algorithm, we minimize a submodular function which can be performed in strongly polynomial time [32, 35] although the best known complexity is $O(n^5 \eta + n^6)$ where $\eta$ is the cost of a function evaluation. Algorithm 1 is guaranteed to converge to a local minima and moreover the algorithm monotonically decreases the function objective at every iteration, as we show below.

**Lemma 3.3.** *[30] Algorithm 1 is guaranteed to decrease the objective function at every iteration. Further, the algorithm is guaranteed to converge to a local minima by checking at most $O(n)$ permutations at every iteration.*

Due to space constraints, we omit the proof of this lemma which is in any case described in [30, 13].

Algorithm 1 requires performing a submodular function minimization at every iteration which while polynomial in $n$ is (due to the complexity described above) not practical for large problem sizes. So while the algorithm reaches a local minima, it can be costly to find it. A desirable result, therefore, would be to develop new algorithms for minimizing DS functions, where the new algorithms have the same properties as the SubSup procedure but are much faster in practice. We give this in the following sections.

can be expressed as a DS function using suitable submodular functions as shown below. The result was first shown in [30] using the Lovász extension. We here give a new combinatorial proof, which avoids Hessians of polyhedral convex functions and which provides a way of constructing (a non-unique) pair of submodular functions $f$ and $g$ for an arbitrary set function $v$.

**Lemma 3.1.** *[30] Given any set function $v$, it can be expressed as a DS functions $v(X) = f(X) - g(X), \forall X \subseteq V$ for some submodular functions $f$ and $g$.*

*Proof.* Given a set function $v$, we can define $\alpha = \min_{X \subset Y \subseteq V \setminus j} v(j|X) - v(j|Y)$[1]. Clearly $\alpha < 0$, since otherwise $v$ would be submodular. Now consider any (strictly) submodular function $g$, i.e., one having $\beta = \min_{X \subset Y \subseteq V \setminus j} g(j|X) - g(j|Y) > 0$. Define $f'(X) = v(X) + \frac{|\alpha'|}{\beta} g(X)$ with any $\alpha' \leq \alpha$. Now it is easy to see that $f'$ is submodular since $\min_{X \subset Y \subseteq V \setminus j} f'(j|X) - f'(j|Y) \geq \alpha + |\alpha'| \geq 0$. Hence $v(X) = f'(X) - \frac{|\alpha'|}{\beta} g(X)$, is a difference between two submodular functions. □

The above proof requires the computation of $\alpha$ and $\beta$ which has, in general, exponential complexity. Using the construction above, however, it is easy to find the decomposition $f$ and $g$ under certain conditions on $v$.

**Lemma 3.2.** *If $\alpha$ or at least a lower bound on $\alpha$ for any set function $v$ can be computed in polynomial time, functions $f$ and $g$ corresponding to $v$ can obtained in polynomial time.*

*Proof.* Define $g$ as $g(X) = \sqrt{|X|}$. Then $\beta = \min_{X \subset Y \subseteq V \setminus j} \sqrt{|X|+1} - \sqrt{|X|} - \sqrt{|Y|+1} + \sqrt{|Y|} = \min_{X \subset V \setminus j} \sqrt{|X|+1} - \sqrt{|X|} - \sqrt{|X|+2} + \sqrt{|X|+1} = 2\sqrt{n-1} - \sqrt{n} - \sqrt{n-2}$. The last inequality follows since the smallest difference in gains will occur at $|X| = n-2$. Hence $\beta$ is easily computed, and given a lower bound on $\alpha$, from Lemma 3.1 the decomposition can be obtained in polynomial time. A similar argument holds for $g$ being other concave functions over $|X|$. □

## 4 Alternate algorithms for minimizing DS functions

In this section we propose two new algorithms to minimize DS functions, both of which are guaranteed to monotonically reduce the objective at every iteration and converge to local minima. We briefly describe these algorithms in the subsections below.

### 4.1 The supermodular-submodular (SupSub) procedure

In the submodular-supermodular procedure we iteratively minimized $f(X) - g(X)$ by replacing $g$ by it's modular lower bound at every iteration. We can instead replace $f$ by it's modular upper bound as is done in Algorithm 2, which leads to the *supermodular-submodular* procedure.

In the SupSub procedure, at every step we perform submodular maximization which, although NP complete to solve exactly, admits a number of fast constant factor approximation algorithms [7, 8]. Notice that we have two modular upper bounds and hence there are a number of ways we can choose between them. One way is to run both maximization procedures with the two modular upper bounds at every iteration in parallel,

---
[1] We denote $j, X, Y : X \subset Y \subseteq V \setminus \{j\}$ by $X \subset Y \subseteq V \setminus j$.

**Algorithm 2** The supermodular-submodular (SupSub) procedure

1: $X^0 = \emptyset$ ; $t \leftarrow 0$ ;
2: **while** not converged (i.e., $(X^{t+1} \neq X^t)$) **do**
3:     $X^{t+1} := \mathrm{argmin}_X \, m^f_{X^t}(X) - g(X)$
4:     $t \leftarrow t+1$
5: **end while**

and choose the one which is better. Here by better we mean the one in which the function value is lesser. Alternatively we can alternate between the two modular upper bounds by first maximizing the expression using the first modular upper bound, and then maximize the expression using the second modular upper bound. Notice that since we perform approximate submodular maximization at every iteration, we are not guaranteed to monotonically reduce the objective value at every iteration. If, however, we ensure that at every iteration we take the next step only if the objective $v$ does not increase, we will restore monotonicity at every iteration. Also, in some cases we converge to local optima as shown in the following theorem.

**Theorem 4.1.** *Both variants of the supermodular-submodular procedure (Algorithm 2) monotonically reduces the objective value at every iteration. Moreover, assuming a submodular maximization procedure in line 3 that reaches a local maxima of $m^f_{X^t}(X) - g(X)$, then if Algorithm 2 does not improve under both modular upper bounds then it reaches a local optima of $v$.*

*Proof.* For either modular upper bound, we have:

$$f(X^{t+1}) - g(X^{t+1}) \overset{a}{\leq} m^f_{X^t}(X^{t+1}) - g(X^{t+1})$$
$$\overset{b}{\leq} m^f_{X^t}(X^t) - g(X^t)$$
$$\overset{c}{=} f(X^t) - g(X^t),$$

where (a) follows since $f(X^{t+1}) \leq m^f_{X^t}(X^{t+1})$, and (b) follows since we assume that we take the next step only if the objective value does not increase and (c) follows since $m^f_{X^t}(X^t) = f(X^t)$ from the tightness of the modular upper bound.

To show that this algorithm converges to a local minima, we assume that the submodular maximization procedure in line 3 converges to a local maxima. Then observe that if the objective value does not decrease in an iteration under both upper bounds, it implies that $m^f_{X^t}(X^t) - g(X^t)$ is already a local optimum in that (for both upper bounds) we have $m^f_{X^t}(X^t \cup j) - g(X^t \cup j) \geq m^f_{X^t}(X^t) - g(X^t), \forall j \notin X^t$ and $m^f_{X^t}(X^t \setminus j) - g(X^t \setminus j) \geq m^f_{X^t}(X^t) - g(X^t), \forall j \in X^t$. Note that $m^f_{X^t,1}(X^t \setminus j) = f(X^t) - f(j|X^t \setminus j) = f(X^t \setminus j)$ and $m^f_{X^t,2}(X^t \cup j) = f(X^t) + f(j|X^t) = f(X^t \cup j)$ and hence if both modular upper bounds are at a local optima, it implies $f(X^t) - g(X^t) = m^f_{X^t,1}(X^t) - g(X^t) \leq m^f_{X^t,1}(X^t \setminus j) - g(X^t \setminus j) = f(X^t \setminus j) - g(X^t \setminus j)$. Similarly $f(X^t) - g(X^t) = m^f_{X^t,2}(X^t) - g(X^t) \leq m^f_{X^t,2}(X^t \cup j) - g(X^t \cup j) = f(X^t \cup j) - g(X^t \cup j)$. Hence $X^t$ is a local optima for $v(X) = f(X) - g(X)$, since $v(X^t) \leq v(X^t \cup j)$ and $v(X^t) \leq v(X^t \setminus j)$. □

To ensure that we take the largest step at each iteration, we can use the recently proposed tight (1/2)-approximation algorithm in [7] for unconstrained non-monotone submodular function maximization — this is the best possible in polynomial time for the class of submodular functions independent of the P=NP question. The algorithm is a form of bi-directional randomized greedy procedure and, most importantly for practical considerations, is linear time [7]. Lastly, note that this algorithm is closely related to a local search heuristic for submodular maximization [8]. In particular, if instead of using the greedy algorithm entirely at every iteration, we take only one local step, we get a local search heuristic. Hence, via the SupSub procedure, we may take larger steps at every iteration as compared to a local search heuristic.

### 4.2 The modular-modular (ModMod) procedure

The submodular-supermodular procedure and the supermodular-submodular procedure were obtained by replacing $g$ by it's modular lower bound and $f$ by it's modular upper bound respectively. We can however replace both of them by their respective modular bounds, as is done in Algorithm 3.

**Algorithm 3** Modular-Modular (ModMod) procedure

1: $X^0 = \emptyset$; $t \leftarrow 0$ ;
2: **while** not converged (i.e., $(X^{t+1} \neq X^t)$) **do**
3:     Choose a permutation $\sigma^t$ whose chain contains the set $X^t$.
4:     $X^{t+1} := \mathrm{argmin}_X \, m^f_{X^t}(X) - h^g_{X^t,\sigma^t}(X)$
5:     $t \leftarrow t+1$
6: **end while**

In this algorithm at every iteration we minimize only a modular function which can be done in $O(n)$ time, so this is extremely easy (i.e., select all negative elements for the smallest minimum, or all non-positive elements for the largest minimum). Like before, since we have two modular upper bounds, we can use any of the variants discussed in the subsection above. Moreover, we are still guaranteed to monotonically decrease the objective at every iteration and converge to a local minima.

**Theorem 4.2.** *Algorithm 3 monotonically decreases the function value at every iteration. If the function value does not increase on checking $O(n)$ different permutations with different elements at adjacent positions and with both modular upper bounds, then we have reached a local minima of $v$.*

*Proof.* Again we can use similar reasoning as the earlier proofs and observe that:

$$f(X^{t+1}) - g(X^{t+1}) \leq m^f_{X^t}(X^{t+1}) - h^g_{X^t,\sigma^t}(X^{t+1})$$
$$\leq m^f_{X^t}(X^t) - h^g_{X^t,\sigma^t}(X^t)$$
$$= f(X^t) - g(X^t)$$

We see that considering $O(n)$ permutations each with different elements at $\sigma^t(|X^t| - 1)$ and $\sigma^t(|X^t| + 1)$, we essentially consider all choices of $g(X^t \cup j)$ and $g(X^t \backslash j)$, since $h^g_{X^t,\sigma^t}(S_{|X^t|+1}) = f(S_{|X^t|+1})$ and $h^g_{X^t,\sigma^t}(S_{|X^t|-1}) = f(S_{|X^t|-1})$. Since we consider both modular upper bounds, we correspondingly consider every choice of $f(X^t \cup j)$ and $f(X^t \backslash j)$. Note that at convergence we have that $m^f_{X^t}(X^t) - h^g_{X^t,\sigma^t}(X^t) \leq m^f_{X^t}(X) - h^g_{X^t,\sigma^t}(X), \forall X \subseteq V$ for $O(n)$ different permutations and both modular upper bounds. Correspondingly we are guaranteed that (since the expression is modular) $\forall j \notin X^t, v(j|X^t) \geq 0$ and $\forall j \in X^t, v(j|X^t \backslash j) \geq 0$, where $v(X) = f(X) - g(X)$. Hence the algorithm converges to a local minima. $\square$

An important question is the choice of the permutation $\sigma^t$ at every iteration $X^t$. We observe experimentally that the quality of the algorithm depends strongly on the choice of permutation. Observe that $f(X) - g(X) \leq m^f_{X^t}(X) - h^g_{X^t,\sigma^t}(X)$, and $f(X^t) - g(X^t) = m^f_{X^t}(X^t) - h^g_{X^t,\sigma^t}(X^t)$. Hence, we might obtain the greatest local reduction in the value of $v$ by choosing permutation $\sigma^* \in \operatorname{argmin}_\sigma \min_X (m^f_{X^t}(X) - h^g_{X^t,\sigma^t}(X))$, or the one which maximizes $h^g_{X^t,\sigma^t}(X)$. We in fact might expect that choosing $\sigma^t$ ordered according to greatest gains of $g$, with respect to $X^t$, we would achieve greater descent at every iteration. Another choice is to choose the permutation $\sigma$ based on the ordering of gains of $v$ (or even $m^f_{X^t}$). Through the former we are guaranteed to at least progress as much as the local search heuristic. Indeed, we observe in practice that the first two of these heuristics performs much better than a random permutation for both the ModMod and the SubSup procedure, thus addressing a question raised in [30] about which ordering to use. Practically for the feature selection problem, the second heuristic seems to work the best.

### 4.3 Constrained minimization of a difference between submodular functions

In this section we consider the problem of minimizing the difference between submodular functions subject to constraints. We first note that the problem of minimizing a submodular function under even simple cardinality constraints in NP hard and also hard to approximate [36]. Since there does not yet seem to be a reasonable algorithm for constrained submodular minimization at every iteration, it is unclear how we would use Algorithm 1. However the problem of submodular maximization under cardinality, matroid, and knapsack constraints though NP hard admits a number of constant factor approximation algorithms [31, 26] and correspondingly the cardinality constraints can be easily introduced in Algorithm 2. Moreover, since a non-negative modular function can be easily, directly and even exactly optimized under cardinality, knapsack and matroid constraints [16], Algorithm 3 can also easily be utilized. In addition, since problems such as finding the minimum weight spanning tree, min-cut in a graph, etc., are polynomial time algorithms in a number of cases, Algorithm 3 can be used when minimizing a non-negative function $v$ expressible as a difference between submodular functions under combinatorial constraints. If $v$ is non-negative, then so is its modular upper bound, and then the ModMod procedure can directly be used for this problem — each iteration minimizes a non-negative modular function subject to combinatorial constraints which is easy in many cases [16, 14].

## 5 Theoretical results

In this section we analyze the computational and approximation bounds for this problem. For simplicity we assume that the function $v$ is normalized, i.e $v(\emptyset) = 0$. Hence we assume that $v$ achieves it minima at a negative value and correspondingly the approximation factor in this case will be less than 1.

We note in passing that the results in this section are mostly negative, in that they demonstrate theoretically how complex a general problem such as $\min_X [f(X) - g(X)]$ is, even for submodular $f$ and $g$. In this paper, rather than consider these hardness results pessimistically, we think of them as providing justification for the heuristic procedures given in Section 4 and [30]. In many cases, inspired heuristics can yield good quality and hence practically useful algorithms for real-world problems. For example, the ModMod procedure (Algorithm 3) and even the SupSub procedure (Algorithm 2) can scale to very large problem sizes, and thus can provide useful new strategies for the applications listed in Section 1.

## 5.1 Hardness

Observe that the class of DS functions is essentially the class of general set functions, and hence the problem of finding optimal solutions is NP-hard. This is not surprising since general set function minimization is inapproximable and there exist a large class of functions where all (adaptive, possibly randomized) algorithms perform arbitrarily poorly in polynomial time [37]. Clearly as is evident from Theorem 3.1, even the problem of finding the submodular functions $f$ and $g$ requires exponential complexity. We moreover show in the following theorem, however, that this problem is multiplicatively inapproximable even when the functions $f$ and $g$ are easy to find.

**Theorem 5.1.** *Unless $P = NP$, there cannot exist any polynomial time approximation algorithm for $\min_X v(X)$ where $v(X) = [f(X) - g(X)]$ is a positive set function and $f$ and $g$ are given submodular functions. In particular, let $n$ be the size of the problem instance, and $\alpha(n) > 0$ be any positive polynomial time computable function of $n$. If there exists a polynomial-time algorithm which is guaranteed to find a set $X' : f(X') - g(X') < \alpha(n)OPT$, where $OPT = \min_X f(X) - g(X)$, then $P = NP$.*

The proof of this theorem is in [13]. In fact we show below that independent of the $P = NP$ question, there cannot exist a sub-exponential time algorithm for this problem. The theorem below gives information theoretic hardness for this problem.

**Theorem 5.2.** *For any $0 < \epsilon < 1$, there cannot exist any deterministic (or possibly randomized) algorithm for $\min_X[f(X) - g(X)]$ (where $f$ and $g$ are given submodular functions), that always finds a solution which is at most $\frac{1}{\epsilon}$ times the optimal, in fewer than $e^{\epsilon^2 n/8}$ queries.*

Again the proof of this theorem is in [13]. Essentially the theorems above say that even when we are given (or can easily find) a decomposition such that $v(X) = f(X) - g(X)$, there exist set functions such that any algorithm (either adaptive or randomized) will perform arbitrarily poorly and this problem is inapproximable. Hence any algorithm trying to find the global optimum for this problem [3] can only be exponential.

## 5.2 Polynomial time lower and upper bounds

The decomposition theorem of [5] shows that any submodular function can be decomposed into a modular function plus a monotone non-decreasing and *totally normalized* polymatroid rank function. Specifically, given submodular $f, g$ we have $f'(X) \triangleq f(X) - \sum_{j \in X} f(j|V \backslash j)$ and $g'(X) \triangleq g(X) - \sum_{j \in X} g(j|V \backslash j)$ with $f', g'$ being totally normalized polymatroid rank functions. Hence we have: $v(X) = f'(X) - g'(X) + k(X)$, with modular $k(X) = \sum_{j \in X} v(j|V \backslash j)$.

The algorithms in the previous sections are all based on repeatedly finding upper bounds for $v$. The following lower bounds directly follow from the results above. (The proof of this is in [13])

**Theorem 5.3.** *We have the following two lower bounds on the minimizers of $v(X) = f(X) - g(X)$:*

$$\min_X v(X) \geq \min_X f'(X) + k(X) - g'(V)$$

$$\min_X v(X) \geq f'(\emptyset) - g'(V) + \sum_{j \in V} \min(k(j), 0)$$

The above lower bounds essentially provide bounds on the minima of the objective and thus can be used to obtain an additive approximation guarantee. The algorithms described in this paper are all polynomial time algorithms (as we show below) and correspondingly from the bounds above we can get an estimate on how far we are from the optimal.

## 5.3 Computational Bounds

We now provide computational bounds for $\epsilon$-approximate versions of our algorithms. Note that this was left as an open question in [30]. Finding the local minimizer of DS functions is PLS complete since it generalizes the problem of finding the local optimum of the MAX-CUT problem [34]. However we show that an $\epsilon$-approximate version of this algorithm will converge in polynomial time.

**Definition 5.1.** *An $\epsilon$-approximate version of an iterative monotone non-decreasing algorithm for minimizing a set function $v$ is defined as a version of that algorithm, where we proceed to step $t + 1$ only if $v(X^{t+1}) \leq v(X^t)(1 + \epsilon)$.*

Note that the $\epsilon$-approximate versions of algorithms 1, 2 and 3, are guaranteed to converge to $\epsilon$-approximate local optima. W.l.o.g., assume that $X^0 = \emptyset$. Then we have the following computational bounds:

**Theorem 5.4.** *The $\epsilon$-approximate versions of algorithms 1, 2 and 3 have a worst case complexity of $O(\frac{\log(|M|/|m|)}{\epsilon}T)$, where $M = f'(\emptyset) + \sum_{j \in V} \min(v(j|V \backslash j), 0) - g'(V)$, $m = v(X^1)$ and $O(T)$ is the complexity of every iteration of the algorithm (which corresponds to respectively the submodular minimization, maximization, or modular minimization in algorithms 1, 2 and 3)..*

The proof of this theorem is in [13]

Observe that for the algorithms we use, $O(T)$ is strongly polynomial in $n$. The best strongly polynomial time algorithm for submodular function minimization is

$O(n^5\eta + n^6)$ [32] (the lower bound is currently unknown). Further the worst case complexity of the greedy algorithm for maximization is $O(n^2)$ while the complexity of modular minimization is just $O(n)$. Note finally that these are worst case complexities and actually the algorithms run much faster in practice.

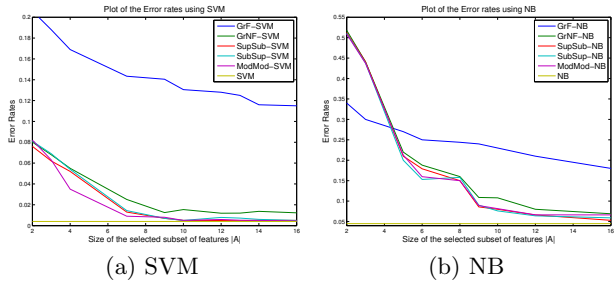

(a) SVM  (b) NB

Figure 1: Plot showing the accuracy rates vs. the number of features on the Mushroom data set.

## 6  Experiments

We test our algorithms on the feature subset selection problem in the supervised setting. Given a set of features $X_V = \{X_1, X_2, \cdots, X_{|V|}\}$, we try to find a subset of these features $A$ which has the most information from the original set $X_V$ about a class variable $C$ under constraints on the size or cost of $A$. Normally the number of features $|V|$ is quite large and thus the training and testing time depend on $|V|$. In many cases, however, there is a strong correlation amongst features and not every feature is novel. We can thus perform training and testing with a much smaller number of features $|A|$ while obtaining (almost) the same error rates.

The question is how to find the most representative set of features $A$. The mutual information between the chosen set of features and the target class $C$, $I(X_A; C)$, captures the relevance of the chosen subset of features. In most cases the selected features are not independent given the class $C$ so the naïve Bayes assumption is not applicable, meaning this is not a pure submodular optimization problem. As mentioned in Section 1, $I(X_A; C)$ can be exactly expressed as a difference between submodular functions $H(X_A)$ and $H(X_A|C)$.

### 6.1  Modular Cost Feature Selection

In this subsection, we look at the problem of maximizing $I(X_A; C) - \lambda|A|$, as a regularized feature subset selection problem. Note that a mutual information $I(X_A; C)$ query can easily be estimated from the data by just a single sweep through this data. Further we have observed that using techniques such as Laplace smoothing helps to improve mutual information estimates without increasing computation. In these experiments, therefore, we estimate the mutual information directly from the data and run our algorithms to find the representative subset of features.

We compare our algorithms on two data sets, i.e., the Mushroom data set [12] and the Adult data set [20] obtained from [9]. The Mushroom data set has 8124 examples with 112 features, while the Adult data set has 32,561 examples with 123 features. In our experiments we considered subsets of features of sizes between 5%-20% of the total number of features by varying $\lambda$. We tested the following algorithms for the feature subset selection problem. We considered two formulations of the mutual information, one under naïve Bayes, where the conditional entropy $H(X_A|C)$ can be written as $H(X_A|C) = \sum_{j \in A} H(X_i|C)$ and another where we do not assume such factorization. We call these two formulations *factored* and *non-factored* respectively. We then considered the simple greedy algorithm, of iteratively adding features at every step to the factored and non-factored mutual information, which we call GrF and GrNF respectively. Lastly, we use the new algorithms presented in this paper on the non-factored mutual information.

We then compare the results of the greedy algorithms with those of the three algorithms for this problem, using two pattern classifiers based on either a linear kernel SVM (using [4]) or a naïve Bayes (NB) classifier. We call the results obtained from the supermodular-submodular heuristic as "SupSub", the submodular-supermodular procedure [30] as "SubSup", and the modular-modular objective as "ModMod." In the SubSup procedure, we use the minimum norm point algorithm [11] for submodular minimization, and in the SubSup procedure, we use the optimal algorithm of [7] for submodular maximization. We observed that the three heuristics generally outperformed the two greedy procedures, and also that GRF can perform quite poorly, thus justifying our claim that the naïve Bayes assumption can be quite poor. This also shows that although the greedy algorithm in that case is optimal, the features are correlated given the class and hence modeling it as a difference between submodular functions gives the best results. We also observed that the SupSub and ModMod procedures perform comparably to the SubSup procedure, while the SubSup procedure is *much* slower in practice. Comparing the running times, the ModMod and the SupSub procedure are each a few times slower then the greedy algorithm (ModMod is slower due computing the modular semigradients), while the SubSup procedure is around 100 times slower. The SubSup procedure is slower due to general submodular function minimization which can be quite slow.

The results for the Mushroom data set are shown in

Figure 1. We performed a 10 fold cross-validation on the entire data set and observed that when using all the features SVM gave an accuracy rate of 99.6% while the all-feature NB model had an accuracy rate of 95.5%. The results for the Adult database are in Figure 2. In this case with the entire set of features the accuracy rate of SVM on this data set is 83.9% and NB is 82.3%.

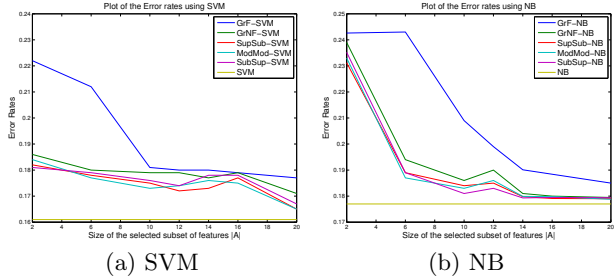

(a) SVM    (b) NB

Figure 2: Plot showing the accuracy rates vs. the number of features on the Adult data set.

In the mushroom data, the SVM classifier significantly outperforms the NB classifier and correspondingly GrF performs much worse than the other algorithms. Also, in most cases the three algorithms outperform GrNF. In the adult data set, both the SVM and NB perform comparably although SVM outperforms NB. However in this case also we observe that our algorithms generally outperform GrF and GrNF.

#### 6.2 Submodular cost feature selection

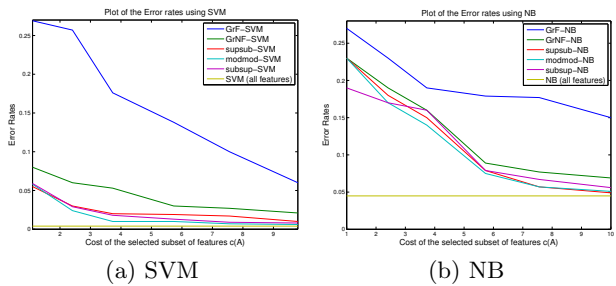

(a) SVM    (b) NB

Figure 3: Plot showing the accuracy rates vs. the cost of features for the Mushroom data set

We perform synthetic experiments for the feature subset selection problem under submodular costs. The cost model we consider is $c(A) = \sum_i \sqrt{m(A \cap S_i)}$. We partitioned $V$ into sets $\{S_i\}_i$ and chose the modular function $m$ randomly. In this set of experiments, we compare the accuracy of the classifiers vs. the *cost* associated with the choice of features for the algorithms. Recall, with simple (modular) cardinality costs the greedy algorithms performed decently in comparison to our algorithms in the adult data set, where the NB assumption is reasonable. However with submodular costs, the objective is no longer submodular even under the NB assumption and thus the greedy algorithms perform much worse. This is unsurprising since the greedy algorithm is approximately optimal only for monotone submodular functions. This is even more strongly evident from the results of the mushrooms data-set (Figure 3)

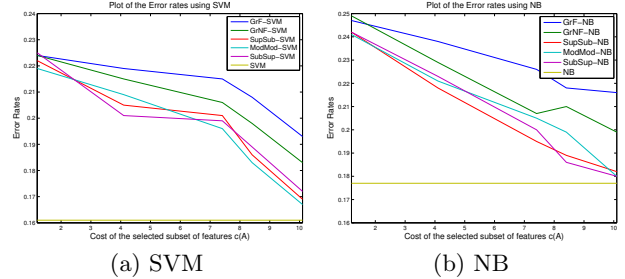

(a) SVM    (b) NB

Figure 4: Plot showing the accuracy rates vs. the cost of features for the Adult data set

## 7 Discussion

We have introduced new algorithms for optimizing the difference between two submodular functions, provided new theoretical understanding that provides some justification for heuristics, have outlined applications that can make use of our procedures, and have tested in the case of feature selection with modular and submodular cost features. Our new ModMod procedure is fast at each iteration and experimentally does about as well as the SupSub and SubSup procedures. The ModMod procedure, moreover, can also be used under various combinatorial constraints, and therefore the ModMod procedure may hold the greatest promise as a practical heuristic. An alternative approach, not yet evaluated, would be to try the convex-concave procedure [39] on the Lovász extensions of $f$ and $g$ since subgradients in such case are so easy to obtain.

**Acknowledgments:** We thank Andrew Guillory, Manas Joglekar, Stefanie Jegelka, and the rest of the submodular group at UW for discussions. This material is based upon work supported by the National Science Foundation under Grant No. (IIS-1162606), and is also supported by a Google, a Microsoft, and an Intel research award.

## References

[1] E. Boros and P. L. Hammer. Pseudo-boolean optimization. *Discrete Applied Mathematics*, 123

(1–3):155 – 225, 2002. ISSN 0166-218X. doi: 10.1016/S0166-218X(01)00341-9.

[2] N. Brenner, S.P. Strong, R. Koberle, W. Bialek, and R.R.R. Steveninck. Synergy in a neural code. *Neural Computation*, 12(7):1531–1552, 2000.

[3] K. Byrnes. Maximizing general set functions by submodular decomposition. *Arxiv preprint arXiv:0906.0120*, 2009.

[4] Chih-Chung Chang and Chih-Jen Lin. LIBSVM: A library for support vector machines. *ACM Transactions on Intelligent Systems and Technology*, 2:27:1–27:27, 2011. Software available at http://www.csie.ntu.edu.tw/~cjlin/libsvm.

[5] W. H. Cunningham. Decomposition of submodular functions. *Combinatorica*, 3(1):53–68, 1983.

[6] J. Edmonds. Submodular functions, matroids and certain polyhedra. *Combinatorial structures and their Applications*, 1970.

[7] M. Feldman, J. Naor, and R. Schwartz. A tight (1/2) linear-time approximation to unconstrained submodular maximization. *To appear, FOCS*, 2012.

[8] Uriel Fiege, Vahab Mirrokni, and Jan Vondrak. Maximizing non-monotone submodular functions. *SIAM J. COMPUT.*, 40(4):1133–1155, July 2007.

[9] A. Frank and A. Asuncion. UCI machine learning repository, 2010. URL http://archive.ics.uci.edu/ml.

[10] S. Fujishige. *Submodular functions and optimization*, volume 58. Elsevier Science, 2005.

[11] S. Fujishige and S. Isotani. A submodular function minimization algorithm based on the minimum-norm base. *Pacific Journal of Optimization*, 7:3–17, 2011.

[12] W. Iba, J. Wogulis, and P. Langley. Trading off simplicity and coverage in incremental concept learning. In *Proceedings of Fifth International Conference on Machine Learning*, pages 73–79, 1988.

[13] R. Iyer and J. Bilmes. Algorithms for approximate minimization of the difference between submodular functions, with applications. *Extended version*, 2012.

[14] S. Jegelka and J. Bilmes. Cooperative cuts: Graph cuts with submodular edge weights. Technical report, Technical Report TR-189, Max Planck Institute for Biological Cybernetics, 2010.

[15] S. Jegelka and J. Bilmes. Submodularity beyond submodular energies: coupling edges in graph cuts. In *Computer Vision and Pattern Recognition (CVPR)*, Colorado Springs, CO, June 2011.

[16] S. Jegelka and J. Bilmes. Online algorithms for submodular minimization with combinatorial constraints. In *Proc. ICML*, 2011.

[17] Stefanie Jegelka and Jeff A. Bilmes. Approximation bounds for inference using cooperative cuts. In *ICML*, Bellevue, Washington, 2011.

[18] Stefanie Jegelka and Jeff A. Bilmes. Submodularity beyond submodular energies: coupling edges in graph cuts. In *Computer Vision and Pattern Recognition (CVPR)*, Colorado Springs, CO, June 2011.

[19] D. Kempe, J. Kleinberg, and E. Tardos. Maximizing the spread of influence through a social network. In *Proc. 9th ACM SIGKDD Intl. Conf. on Knowledge Discovery and Data Mining*, 2003.

[20] R. Kohavi. Scaling up the accuracy of naive-bayes classifiers: A decision-tree hybrid. In *Proceedings of the second international conference on knowledge discovery and data mining*, volume 7, 1996.

[21] V. Kolmogorov and C. Rother. Minimizing non-submodular functions with graph cuts–a review. *IEEE TPAMI*, 29(7):1274–1279, 2007.

[22] V. Kolmogorov and R. Zabih. What energy functions can be minimized via graph cuts? *IEEE TPAMI*, 26(2):147–159, 2004.

[23] A. Krause and C. Guestrin. Near-optimal nonmyopic value of information in graphical models. In *Proceedings of Uncertainity in Artificial Intelligence*. UAI, 2005.

[24] A. Krause, A. Singh, and C. Guestrin. Near-optimal sensor placements in gaussian processes: Theory, efficient algorithms and empirical studies. *The Journal of Machine Learning Research*, 9:235–284, 2008.

[25] Andreas Krause, Brendan McMahan, Carlos Guestrin, and Anupam Gupta. Robust submodular observation selection. *Journal of Machine Learning Research (JMLR)*, 9:2761–2801, December 2008.

[26] J. Lee, V.S. Mirrokni, V. Nagarajan, and M. Sviridenko. Non-monotone submodular maximization under matroid and knapsack constraints. In *Proceedings of the 41st annual ACM symposium on Theory of computing*, pages 323–332. ACM, 2009.

[27] Hui Lin and Jeff Bilmes. Multi-document summarization via budgeted maximization of submodular functions. *NAACL*, 2010.

[28] Hui Lin and Jeff Bilmes. A class of submodular functions for document summarization. *In Proc. ACL*, 2011.

[29] Hui Lin and Jeff A. Bilmes. Optimal selection of limited vocabulary speech corpora. In *Proc.*

*Annual Conference of the International Speech Communication Association (INTERSPEECH)*, Florence, Italy, August 2011.

[30] Mukund Narasimhan and Jeff Bilmes. A submodular-supermodular procedure with applications to discriminative structure learning. In *Uncertainty in Artificial Intelligence (UAI)*, Edinburgh, Scotland, July 2005. Morgan Kaufmann Publishers.

[31] G.L. Nemhauser, L.A. Wolsey, and M.L. Fisher. An analysis of approximations for maximizing submodular set functions—i. *Mathematical Programming*, 14(1):265–294, 1978.

[32] J.B. Orlin. A faster strongly polynomial time algorithm for submodular function minimization. *Mathematical Programming*, 118(2):237–251, 2009.

[33] J. Pearl. *Probabilistic Reasoning in Intelligent Systems: Networks of Plausible Inference*. Morgan Kaufmann, 2nd printing edition, 1988.

[34] A.A. Schäffer. Simple local search problems that are hard to solve. *SIAM journal on Computing*, 20:56, 1991.

[35] A. Schrijver. A combinatorial algorithm minimizing submodular functions in strongly polynomial time. *Journal of Combinatorial Theory, Series B*, 80(2):346–355, 2000.

[36] Z. Svitkina and L. Fleischer. Submodular approximation: Sampling-based algorithms and lower bounds. In *Foundations of Computer Science, 2008. FOCS'08. IEEE 49th Annual IEEE Symposium on*, pages 697–706. IEEE, 2008.

[37] L. Trevisan. Inapproximability of combinatorial optimization problems. *The Computing Research Repository*, 2004.

[38] M.J. Wainwright and M.I. Jordan. Graphical models, exponential families, and variational inference. *Foundations and Trends® in Machine Learning*, 1(1-2):1–305, 2008.

[39] A.L. Yuille and A. Rangarajan. The concave-convex procedure (cccp). *Advances in Neural Information Processing Systems*, 2:1033–1040, 2002.